\documentclass[a4paper]{article}

\usepackage{INTERSPEECH2016}

\usepackage[utf8]{inputenc}
\usepackage{graphicx}
\usepackage{amssymb,amsmath,bm}
\usepackage{textcomp}

\usepackage[T1]{fontenc}
\usepackage{microtype}
\usepackage{tcolorbox}
\tcbuselibrary{listings}
\usepackage{inconsolata}
\usepackage{epstopdf}
\usepackage{tikz}
\usepackage[hidelinks]{hyperref}
\usetikzlibrary{arrows}

\newtcblisting{code}{%
  listing options={
    basicstyle=\small\ttfamily,
    breaklines=true,
    columns=fullflexible
  },
  colframe=black!50!black,
  hbox,
  listing only,
  arc=0pt,
  outer arc=0pt,
  top=1.5mm,
  bottom=1.5mm,
  left=3mm,
  right=3mm,
  boxrule=1.0pt,
  colback=yellow!20
}

\ninept

\title{TheanoLM --- An Extensible Toolkit for Neural Network Language Modeling}

\makeatletter
\def\name#1{\gdef\@name{#1\\}}
\makeatother \name{{\em Seppo Enarvi, Mikko Kurimo}}

\address{Aalto University, Finland \\
  {\small \tt firstname.lastname@aalto.fi}
}

\begin{document}

\maketitle

\begin{abstract}
We present a new tool for training neural network language models (NNLMs),
scoring sentences, and generating text. The tool has been written using Python
library Theano, which allows researcher to easily extend it and tune any aspect
of the training process. Regardless of the flexibility, Theano is able to
generate extremely fast native code that can utilize a GPU or multiple CPU cores
in order to parallelize the heavy numerical computations. The tool has been
evaluated in difficult Finnish and English conversational speech recognition
tasks, and significant improvement was obtained over our best back-off n-gram
models. The results that we obtained in the Finnish task were compared to those
from existing RNNLM and RWTHLM toolkits, and found to be as good or better,
while training times were an order of magnitude shorter.
\end{abstract}


\section{Introduction}

Neural network language models (NNLM) are known to outperform traditional n-gram
language models in speech recognition accuracy \cite{Schwenk:2002,
Mikolov:2012}. For modeling word sequences with temporal dependencies, the
recurrent neural network (RNN) is an attractive model as it is not limited to a
fixed window size. Perhaps the simplest variation of RNN that has been used for
language modeling contains just one hidden layer \cite{Mikolov:2010}. The
ability of an RNN to model temporal dependencies is limited by the vanishing
gradient problem. Various modifications have been proposed to the standard RNN
structure that reduce this problem, such as the popular long short-term memory
(LSTM) \cite{Hochreiter:1997}.

Figure \ref{fig:lstm-network} shows a typical LSTM network. Following the
architecture by Bengio et al \cite{Bengio:2003}, the first layer projects the
words $w_t$ into a continuous lower-dimensional vector space, which is followed
by a hidden layer. In recurrent networks the hidden layer state $h_t$ is passed
to the next time step. LSTM is a special case of a recurrent layer that also
passes \emph{cell state} $C_t$. Sigmoid gates control what information will be
added to or removed from the cell state, making it easy to maintain important
information in the cell state over extended periods of time. The final neural
network layer normalizes the output probabilities using $softmax$.

Another influential design choice concerns performance when the vocabulary is
large. The computational cost of training and evaluating a network with softmax
output layer is highly dependent on the number of output neurons, i.e. the size
of the vocabulary. Feedforward networks are basically n-gram models, so it is
straightforward to use the neural network to predict only words in a short-list
and use a back-off model for the infrequent words \cite{Schwenk:2013}. Replacing
a single softmax layer with a hierarchy of softmax layers \cite{Morin:2005}
improves performance, although the number of model parameters is not reduced.
Using word classes in the input and output of the network
\cite{Sundermeyer:2014} has the benefit that the model becomes smaller, which
may be necessary for using a GPU.

While several toolkits have been made available for language modeling with
neural networks \cite{Schwenk:2013,Sundermeyer:2014,Mikolov:2011:ASRU}, they do
not support research and prototyping work well. The most important limitation is
the difficulty in extending the toolkits with recently proposed methods and
different network architectures. Also, training large networks is very slow
without GPU support. TheanoLM is a new NNLM tool that we have developed,
motivated by our ongoing research on improving conversational Finnish automatic
speech recognition (ASR). Our goal is to make it versatile and fast, easy to
use, and write code that is easily extensible.

In the next section we will give an introduction on how TheanoLM works. Then we
will evaluate it on two conversational ASR tasks with large data sets. In order
to verify that it works correctly, the results will be compared to those from
other toolkits.

\begin{figure}[t]
\centering
\begin{tikzpicture}[
  font=\sffamily\scriptsize,
  every matrix/.style={ampersand replacement=\&,column sep=6mm,row sep=7mm},
  input/.style={draw,thick,circle,fill=blue!20},
  output/.style={inner sep=.1cm},
  layer/.style={draw,thick,fill=yellow!20,inner sep=.3cm},
  arrow/.style={->,>=stealth',shorten >=1pt,semithick},
  arrow/.style={->,>=stealth',shorten >=1pt,semithick},
  every node/.style={align=center}]

  \matrix {
    \node[output] (y0) {$P(w_1 | w_0)$}; \&
    \node[output] (y1) {$P(w_2 | w_1, w_0)$}; \&
    \node[output] (y2) {$P(w_3 | w_2, w_1, w_0)$}; \&
    \node[coordinate] (y3) {}; \\
    
    \node[layer] (output0) {$softmax$}; \&
    \node[layer] (output1) {$softmax$}; \&
    \node[layer] (output2) {$softmax$}; \&
    \node[coordinate] (output3) {}; \\

    \node[layer] (skip0) {$tanh$}; \&
    \node[layer] (skip1) {$tanh$}; \&
    \node[layer] (skip2) {$tanh$}; \&
    \node[coordinate] (skip3) {}; \\

    \node[coordinate] (h0) {}; \&
    \node[coordinate] (h1) {}; \&
    \node[coordinate] (h2) {}; \&
    \node[coordinate] (h3) {}; \\

    \node[layer] (hidden0) {LSTM}; \&
    \node[layer] (hidden1) {LSTM}; \&
    \node[layer] (hidden2) {LSTM}; \&
    \node[coordinate] (hidden3) {}; \\

    \node[layer] (projection0) {projection}; \&
    \node[layer] (projection1) {projection}; \&
    \node[layer] (projection2) {projection}; \&
    \node[coordinate] (projection3) {}; \\

    \node[input] (x0) {$w_0$}; \&
    \node[input] (x1) {$w_1$}; \&
    \node[input] (x2) {$w_2$}; \&
    \node[coordinate] (x3) {}; \\
  };

  \draw[arrow] (x0) -- (projection0);
  \draw[arrow] (x1) -- (projection1);
  \draw[arrow] (x2) -- (projection2);


  \draw[arrow] (projection0) -- (hidden0);
  \draw[arrow] (projection1) -- (hidden1);
  \draw[arrow] (projection2) -- (hidden2);

  \draw[arrow] (hidden0) -- node[midway,above] {$C_0$} (hidden1);
  \draw[arrow] (hidden1) -- node[midway,above] {$C_1$} (hidden2);
  \draw[arrow] (hidden2) -- node[midway,above] {$C_2$} (hidden3);

  \draw[arrow] (hidden0) -- (skip0);
  \draw[arrow] (hidden1) -- (skip1);
  \draw[arrow] (hidden2) -- (skip2);

  \draw[arrow] (h0) -| node[midway,above] {$h_0$} (hidden1.145);
  \draw[arrow] (h1) -| node[midway,above] {$h_1$} (hidden2.145);
  \draw[arrow] (h2) -- node[midway,above] {$h_2$} (h3);

  \draw[arrow] (skip0) -- (output0);
  \draw[arrow] (skip1) -- (output1);
  \draw[arrow] (skip2) -- (output2);

  \draw[arrow] (output0) -- (y0);
  \draw[arrow] (output1) -- (y1);
  \draw[arrow] (output2) -- (y2);

\end{tikzpicture}
\caption{{\it Recurrent NNLM with LSTM and $tanh$ hidden layers. 1-of-N encoded
words $w_t$ are projected into lower-dimensional vectors. An LSTM layer passes
the hidden state $h_t$ to the next time step, like a standard recurrent layer,
but also the cell state $C_t$, which is designed to convey information over
extended time intervals.}}
\label{fig:lstm-network}
\end{figure}
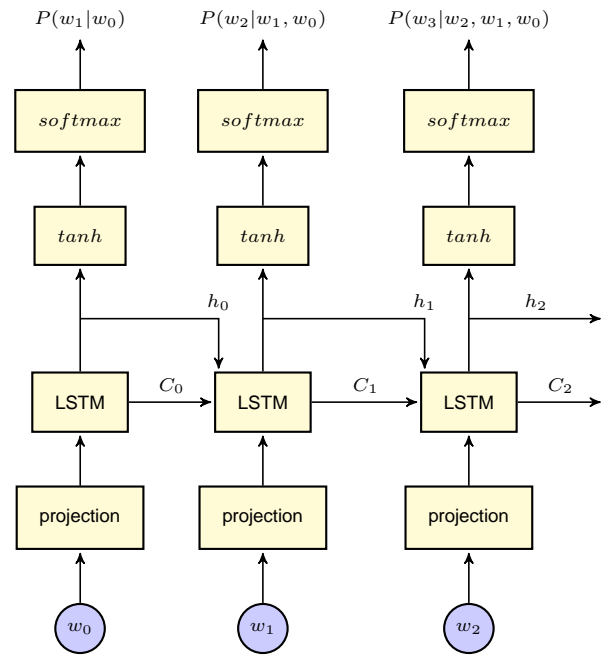

\begin{figure*}[t]
\centering
\begin{code}
input type=class name=class_input
layer type=projection name=projection_layer input=class_input size=500
layer type=dropout name=dropout_layer_1 input=projection_layer dropout_rate=0.25
layer type=lstm name=hidden_layer_1 input=dropout_layer_1 size=1500
layer type=dropout name=dropout_layer_2 input=hidden_layer_1 dropout_rate=0.25
layer type=tanh name=hidden_layer_2 input=dropout_layer_2 size=1500
layer type=dropout name=dropout_layer_3 input=hidden_layer_2 dropout_rate=0.25
layer type=softmax name=output_layer input=dropout_layer_3
\end{code}
\caption{{\it An example of a network architecture description.}}
\label{fig:lstm-network-config}
\end{figure*}

\section{TheanoLM}

Theano is a Python library for high-performance mathematical computation
\cite{Bastien:2012}. It provides a versatile interface for building neural
network applications, and has been used in many neural network tasks. For
language modeling, toolkits such as RNNLM are more popular, because they are
easier to approach. We have developed TheanoLM as a complete package for
training and applying recurrent neural network language models in speech
recognition and machine translation, as well as generating text by sampling from
an NNLM. It is freely available on
GitHub\footnote{\url{https://github.com/senarvi/theanolm}}.

The neural network is represented in Python objects as a symbolic graph of
mathematical expressions. Theano performs symbolic differentiation, making it
easy to implement gradient-based optimization methods. We have already
implemented Nesterov's Accelerated Gradient \cite{Nesterov:1983}, Adagrad
\cite{Duchi:2011}, Adadelta \cite{Zeiler:2012}, Adam \cite{Kingma:2015}, and
RMSProp optimizers, in addition to plain Stochastic Gradient Descent (SGD).

Evaluation of the expressions is performed using native CPU and GPU code
transparently. The compilation does introduce a short delay before the program
can start to train or use a model, but in a typical application the delay is
negligible compared to the actual execution. On the other hand, the execution
can be highly parallelized using a GPU, speeding up training of large networks
to a fraction of CPU training times.

Standard SGD training is very sensitive to the learning rate hyperparameter. The
initial value should be as high as possible, given that the optimization still
converges. Gradually decreasing (annealing) the learning rate enables finer
adjustments later in the optimization process. TheanoLM can perform
cross-validation on development data at regular intervals, in order to decide
how quickly annealing should occur. However, adaptive learning rate methods such
as Adagrad and Adadelta do not require manual tuning of the learning
rate---cross-validation is needed only for determining when to stop training.

Especially in Finnish and other highly agglutinative languages the vocabulary is
too large for the final layer to predict words directly. In this work we use
class-based models, where each word $w_t$ belongs to exactly one word class
$c(w_t)$:

\begin{equation}
P(w_t | w_{t-1} \ldots) = P(c(w_t) | c(w_{t-1}) \ldots) P(w_t | c(w_t))
\end{equation}

When the classes are chosen carefully, this model will not necessarily perform
worse than a word-based model in Finnish data, because there are not enough
examples of the rarer words to give robust estimates of word probabilities.
The advantage of this solution is that the model size depends on the number of
classes instead of the number of words.

An arbitrary network architecture can be provided in a text file as a list of
layer descriptions. The layers have to specified in the order that the network
is constructed, meaning that the network has to be acyclic. The network may
contain word and class inputs, which have to be followed by a projection layer.
A projection layer may be followed by any number of LSTM and GRU \cite{Cho:2014}
layers, as well as non-recurrent layers. The final layer has to be a softmax or
a hierarchical softmax layer. Multiple layers may have the same element in their
input, and a layer may have multiple inputs, in which case the inputs will be
concatenated.

The example in Figure \ref{fig:lstm-network-config} would create an LSTM network
with a projection layer of 500 neurons and two hidden layers of 1500 neurons. 
After each of those layers is a dropout \cite{Srivastava:2014} layer, which
contains no neurons, but only sets some activations randomly to zero at train
time to prevent overfitting. This is the configuration that we have used in this
paper to train the larger TheanoLM models.

LSTM and GRU architectures help avoid the vanishing gradient problem. In order
to prevent gradients from exploding, we limit the norm of the gradients, as
suggested by Mikolov in his thesis \cite{Mikolov:2012}.

\section{Conversational Finnish ASR Experiment}

\subsection{Data}

In order to develop models for conversational Finnish ASR, we have recorded and
transcribed conversations held by students in pairs, in the basic course in
digital signal processing at Aalto University. The collected corpus, called
\emph{Aalto University DSP Course Conversation Corpus (DSPCON)} is available for
research use in the Language Bank of
Finland\footnote{\url{http://urn.fi/urn:nbn:fi:lb-2015101901}}. The corpus is
updated annually; currently it includes recordings from years 2013--2015. In
addition we have used the spontaneous sentences of SPEECON corpus, FinDialogue
subcorpus of FinINTAS\footnote{\url{http://urn.fi/urn:nbn:fi:lb-20140730194}},
and some transcribed radio conversations, totaling 34 hours of acoustic training
data. This is practically all the conversational Finnish data that is available
for research.

We have collected language modeling data from the Internet, and filtered it to
match transcribed conversations \cite{Kurimo:2016}. Similar data is available in
the Language Bank of
Finland\footnote{\url{http://urn.fi/urn:nbn:fi:lb-201412171}}. This was
augmented with 43,000 words of transcribed spoken conversations from DSPCON,
totaling 76 million words. A 8,900-word development set was used in language
modeling.

A set of 541 sentences from unseen speakers, totaling 44 minutes and
representing the more natural spontaneous speech of various topics, was used for
evaluation. Because of the numerous ways in which conversational Finnish can be
written down, ASR output should be evaluated on transcripts that contain such
alternative word forms. As we did not have another suitable evaluation set, the
same data was used for optimizing language model weights. The lattices were
generated using Aalto ASR \cite{Hirsimaki:2009} and a triphone HMM acoustic
model.

\subsection{Models}

In this task we have obtained results also from other freely available NNLM
toolkits, RWTHLM \cite{Sundermeyer:2014} and RNNLM \cite{Mikolov:2011:ASRU}, for
comparison. With RWTHLM and TheanoLM we have used one LSTM layer and one $tanh$
layer on top of the projection layer (100+300+300 neurons), as in Figure
\ref{fig:lstm-network}. RNNLM supports only a simple recurrent network with one
hidden layer. Because of the faster training time with TheanoLM, we were also
able to try a larger model with five times the number of neurons in each layer.
This model includes dropout after each layer. The architecture description is
given in Figure \ref{fig:lstm-network-config}. We trained also models with third
existing toolkit, CSLM \cite{Schwenk:2013}, but were unable to get meaningful
sentence scores. CSLM supports only feedforward networks, but is very fast
because of GPU support.

\begin{table}[t]
\caption{{\it Language model training times and word error rates (\%) given by
the model alone and interpolated with the best Kneser-Ney model. Finnish
results are from the evaluation set, but the same set was used for optimizing
language model weights.}}
\label{tab:wer-results}
\begin{center}
\begin{tabular}{@{}l@{\hskip 0.1cm}c@{\hskip 0.1cm}c@{\hskip 0.1cm}c@{\hskip 0.1cm}c@{\hskip 0.1cm}c@{\hskip 0.1cm}}
\hline
                         & Training & \multicolumn{2}{c}{Dev}
                                    & \multicolumn{2}{c}{Eval} \\
Model                    & time     & Alone & Int. & Alone & Int. \\
\hline
\textbf{Finnish} \\
KN word 4-gram           & 7 min    & 52.1 \\
KN word 4-gram weighted  & 19 min   & 51.0 \\
KN class 4-gram weighted & 55 min   & 51.2 & 50.5 \\
RNNLM 300                & 361 h    & 50.4 & 49.8 \\
RWTHLM 100+300+300       & 480 h \textsuperscript{\textdagger}
                                    & 49.4 & 48.6 \\
TheanoLM 100+300+300     & 20 h     & 49.1 & 49.0 \\
TheanoLM 500+1500+1500   & 139 h    & 48.7 & 48.4
\vspace{2mm} \\
\textbf{English} \\
KN word 4-gram           & 8 min    & 42.1 &      & 41.9 \\
KN word 4-gram weighted  & 15 min   & 41.0 &      & 41.2 \\
TheanoLM 100+300+300     & 62 h     & 40.1 & 39.6 & 41.2 & 40.5 \\
TheanoLM 500+1500+1500   & 290 h    & 39.6 & 39.5 & 40.5 & 40.0 \\
\hline
\end{tabular}
\end{center}
\textsuperscript{\textdagger}\footnotesize{Using 12 CPUs.}
\end{table}

The toolkits also differ in how they handle the vocabulary. With RWTHLM and
TheanoLM we used word classes. 2000 word classes were created using the exchange
algorithm \cite{Kneser:1993}, which tries to optimize the log probability of a
bigram model on the training data, by moving words between classes. RNNLM
creates classes by frequency binning, but uses words in the input and output of
the neural network. Classes are used for decomposition of the output layer,
which speeds up training and evaluation \cite{Mikolov:2011:ICASSP}, but with
millions of words in the vocabulary, the number of parameters in the RNNLM model
with 300 neurons is larger than in the RWTHLM and TheanoLM models with
100+300+300 neurons.

With TheanoLM we have used Adagrad optimizer without annealing. While we could
not evaluate different optimizers extensively, Adagrad seemed to be among the
best in terms of both speed of convergence and performance of the final model.
Nesterov's Accelerated Gradient with manual annealing gave a slightly better
model with considerably longer training time. The other toolkits use standard
SGD.

Kneser-Ney smoothed 4-grams were used in the back-off models. The data sets
collected from different sources varied in size and quality. Instead of pooling
all the data together, the baseline back-off models were weighted mixtures of
models estimated from each subset. The weights were optimized using development
data. The back-off model vocabulary was limited to 200,000 of the 2,400,000
different word forms that occurred in the training data, selected with the EM
algorithm to maximize the likelihood of the development data
\cite{Venkataraman:2003}. Out-of-vocabulary rate in the evaluation data was 5.06
\%.

\subsection{Results}

\begin{table}[t]
\caption{{\it Development and evaluation set perplexities from the
full-vocabulary models.}}
\label{tab:ppl-results}
\begin{center}
\begin{tabular}{lcc}
\hline
      & \multicolumn{2}{c}{Perplexity} \\
Model & Dev & Eval \\
\hline
\multicolumn{3}{l}{\textbf{Finnish}} \\
KN class 4-gram weighted & 755 & 763 \\
RWTHLM 100+300+300       & 687 & 743 \\
RNNLM 300                & 881 & 872 \\
TheanoLM 100+300+300     & 677 & 701 \\
TheanoLM 500+1500+1500   & 609 & 642
\vspace{2mm} \\
\multicolumn{3}{l}{\textbf{English}} \\
KN class 4-gram weighted & 98  & 91 \\
TheanoLM 100+300+300     & 102 & 99 \\
TheanoLM 500+1500+1500   & 90  & 88 \\
\hline
\end{tabular}
\end{center}
\end{table}

Evaluation of neural network probabilities is too slow to be usable in the first
decoding pass of a large-vocabulary continuous speech recognizer. Another pass
is performed by decoding and rescoring word lattices created using a traditional
n-gram model. RWTHLM is able to rescore word lattices directly, the other
toolkits can only rescore n-best lists created from word lattices.

Word error rates (WER) after rescoring are shown in Table~\ref{tab:wer-results}.
The table also includes word error rates given by interpolation of NNLM scores
with the \emph{KN word 4-gram weighted} model. We interpolated the sentence
scores (log probabilities) as

\begin{equation}
\begin{split}
\log P &= (1 - \lambda) s_{bo} \log P_{bo}(w_1 \ldots w_n) \\
       &+ \lambda s_{nn} \log P_{nn}(w_1 \ldots w_{n}).
\end{split}
\end{equation}

$s_{bo}$ is the LM scale factor that was optimal for the back-off model. The
same value was used for generating the n-best lists. $s_{nn}$ and $\lambda$,
the NNLM scale factor and interpolation weight, were optimized from a range of
values.

Vocabulary size affects perplexity computation, so we have omitted the
limited-vocabulary models from the perplexity results in
Table~\ref{tab:ppl-results}. Finnish vocabulary was five times larger, so the
perplexities are considerably higher than in the English language experiment, as
expected. The values are as reported by each tool; there might be differences in
e.g. how unknown words and sentence starts and ends are handled. Perplexities of
the Kneser-Ney models were computed using SRILM, which excludes unknown words
from the computation. With TheanoLM we used the same behaviour
(\textit{-{}-unk-penalty 0}).

We have also recorded the training times in Table~\ref{tab:wer-results},
although it has to be noted that the jobs were run in a compute cluster that
assigns them to different hardware, so the reported durations are only
indicative. RWTHLM supports parallelization through various math libraries.
RNNLM is able to use only one CPU core, which means that the computation is
inevitably slow. TheanoLM was used with an Nvidia Tesla GPU.

When rescoring ASR output, all neural network models outperformed the back-off
models, even without interpolation with back-off scores. Since we get the
back-off model scores from the first decoding pass without additional cost, it
is reasonable to look at the performance when both models are combined with
interpolation. This further improves the results. The back-off model is clearly
improved by weighting individual corpora, because the different corpora are not
homogeneous.  At the time of writing we have implemented training set weighting
schemes in TheanoLM as well, but so far the improvements have been smaller than
with the well-established back-off model weighting.

RWTHLM and RNNLM were stopped after 8 training epochs. RNNLM did not improve the
baseline as much as expected, but training further iterations could have
improved its performance.

\section{English Meeting Recognition Experiment}

For the English task we used more advanced acoustic models trained using Kaldi
\cite{Povey:2011} with discriminative training using the maximum mutual
information (MMI) criterion. The lattices were generated using maximum
likelihood linear regression (MLLR) speaker adaptation. The acoustic training
data was 73 hours from ICSI Meeting
Corpus\footnote{\url{https://catalog.ldc.upenn.edu/LDC2004S02}}. For language
model training we used also part 2 of the Fisher
corpus\footnote{\url{https://catalog.ldc.upenn.edu/LDC2005T19}} and matching
data from the Internet. In total the text data contained 159 million words, of
which 11 million words were transcribed conversations. The development and
evaluation data were from the NIST Rich Transcription 2004 Spring Meeting
Recognition Evaluation. The development set contained 18,000 words and the
evaluation set was 104 minutes and consisted of 2450 utterances.

In the back-off models, vocabulary was limited to 50,000 of the 470,000
different words that occurred in training data. Only 0.30 \% of the evaluation
set tokens were left outside the vocabulary. TheanoLM was used to train models
of the same architecture using the same parameters as in the Finnish task. In
this task we had roughly twice the amount of data. Training was more than two
times slower, but still manageable. The results in Table~\ref{tab:wer-results}
show that the benefit from the larger network is pronounced in this task. The
smaller network is clearly incapable of modeling the larger data set as well.

\section{Conclusions}

Several different NNLM toolkits, implemented in C++, are freely available.
However, the field is rapidly changing, and a toolkit should be easily
extensible with the latest algorithms. Also, different languages and data sets
work best with different algorithms and architectures, so a good solution needs
to be versatile. We offer a new toolkit that has been implemented using Python
and Theano, provides implementations of the latest training methods for
artificial neural networks, and is easily extensible. Another important
advantage is speed---Theano provides highly optimized C++ implementations of the
expensive matrix operations, and can automatically utilize a GPU.

While back-off n-gram models are still two orders of magnitude faster to train,
training a model using TheanoLM was an order of magnitude faster than training a
similar model with the other toolkits. The speed advantage of TheanoLM was
mainly due to GPU support, but the Adagrad optimizer that we used with TheanoLM
was also faster to converge. 4 epochs were required for convergence using
Adagrad, while the other toolkits continued to improve perplexity for at least 8
epochs. The faster training time makes it practical to train larger networks
with TheanoLM. When more data is available, the benefit of a larger network
becomes pronounced, and training without GPU support becomes impractical.

In a conversational Finnish speech recognition task we have used practically all
the data that is available for research. A 4-gram Kneser-Ney model, trained
simply by concatenating the training data, gave us 52.1 \% WER. As a baseline we
took a mixture model that was combined from smaller models with optimized
interpolation weights. The baseline model gave 51.0 \% WER. 48.4 \% WER was
reached when interpolating TheanoLM and baseline LM scores, a relative
improvement of 5.1 \%. RWTHLM gives similar results with a similar neural
network architecture, which increases our confidence in the correctness of the
implementation.

Evaluating the progress we have made in the conversational Finnish task, 48.4 \%
WER is 10.5 \% better than our previous record, 54.1 \% \cite{Kurimo:2016}.
However, we have collected new acoustic data, which explains the 51.0 \%
baseline in this paper. The acoustic models used in these experiments are still
not state of the art. In order to find out the absolute performance of these
models, we will also need to obtain a proper development set for optimizing
language model scale and interpolation weight.

It appears that some types of NNLMs work better with one language than another.
RNNLM did not perform as well as we expected in the Finnish task, probably
because the network architecture was not optimal. RNNLM offers only a simple
network architecture that takes words as input and contains one hidden layer.
The number of input connections is huge with the Finnish vocabulary, and the
size of the hidden layer is limited if we want training time to be reasonable.
Previously interpolating RNNLM with a Kneser-Ney model has been shown to give
3 to 8 \% relative improvement in conversational English ASR
\cite{Kombrink:2011}.

In the English meeting recognition task more data was available, and the smaller
neural network was not better than the weighted mixture back-off model. Still
the neural network brings new information so much that interpolation of the
smaller NNLM scores with back-off scores improves from the baseline of 41.2 \%
WER to 40.5 \%. The larger network brings 2.9 \% improvement to 40.0 \% WER.
When more data is available, a larger network is needed, which makes training
speed even more important.

So far our research focus has been on conversational Finnish and other highly
agglutinative languages. Considering that as much effort has not gone into the
English task, the results are satisfactory, and show that TheanoLM works also in
a relatively standard conversational English task.

\section{Acknowledgements}

This work was financially supported by the Academy of Finland under the grant
number 251170 and made possible by the computational resources provided by the
Aalto Science-IT project.

\bibliographystyle{IEEEtran}
\bibliography{references}

\end{document}